\documentclass{article}

\usepackage{microtype}
\usepackage{graphicx}
\usepackage{subfigure}
\usepackage{booktabs} 
\usepackage{multicol}
\usepackage{multirow}
\usepackage{algorithm}
\usepackage{algorithmic}
\usepackage{amsmath}
\usepackage{hyperref}
\usepackage{url}

\usepackage{array}
\newcolumntype{H}{>{\setbox0=\hbox\bgroup}c<{\egroup}@{}}

\renewcommand{\algorithmicrequire}{ \textbf{Input:}} 
\renewcommand{\algorithmicensure}{ \textbf{Output:}} 

\usepackage[accepted]{icml2024}

\usepackage{amsmath}
\usepackage{amssymb}
\usepackage{mathtools}
\usepackage{amsthm}

\usepackage{mathtools}
\DeclarePairedDelimiter\ceil{\lceil}{\rceil}

\usepackage[capitalize,noabbrev]{cleveref}

\theoremstyle{plain}

\theoremstyle{definition}

\theoremstyle{remark}

\usepackage[textsize=tiny]{todonotes}

\icmltitlerunning{Flash Communication}

\begin{document}

\twocolumn[
\icmltitle{Flash Communication: Reducing Tensor Parallelization Bottleneck for Fast Large Language Model Inference}



\icmlsetsymbol{equal}{*}
\begin{icmlauthorlist}
\icmlauthor{Qingyuan Li}{meituan}
\icmlauthor{Bo Zhang}{meituan}
\icmlauthor{Liang Ye}{meituan}
\icmlauthor{Yifan Zhang}{meituan}
\icmlauthor{Wei Wu}{meituan}
\icmlauthor{Yerui Sun}{meituan}
\icmlauthor{Lin Ma}{meituan}
\icmlauthor{Yuchen Xie}{meituan}
\end{icmlauthorlist}

\icmlaffiliation{meituan}{Meituan}

\icmlcorrespondingauthor{Qingyuan Li}{liqingyuan02@meituan.com}

\icmlkeywords{Machine Learning, ICML}

\vskip 0.3in
]



\printAffiliationsAndNotice{\icmlEqualContribution} 


\begin{abstract}
The ever-increasing sizes of large language models necessitate distributed solutions for fast inference that exploit multi-dimensional parallelism, where computational loads are split across various accelerators such as GPU clusters. However, this approach often introduces significant communication overhead, especially on devices with limited bandwidth. In this paper, we introduce \emph{Flash Communication}, a novel low-bit compression technique designed to alleviate the tensor-parallelism communication bottleneck during inference. Our method substantially boosts intra-node communication speed by more than \textbf{3$\times$} and reduces the \emph{time-to-first-token} by \textbf{2$\times$}, with nearly no sacrifice in model accuracy. Extensive experiments on various up-to-date LLMs demonstrate the effectiveness of our approach.
\end{abstract}

\section{Introduction}

To date, the number of parameters of large language models has tremendously increased. For instance, GPT-3~\cite{brown2020language} has 175B, DeepSeek V2~\cite{liu2024deepseek} utilizes 236B, LLaMA-3 ~\cite{dubey2024llama} reaches 450B. Their enormous sizes create big challenges for both training and inference.

\begin{figure}[ht]
\begin{center}
\centerline{\includegraphics[width=\columnwidth]{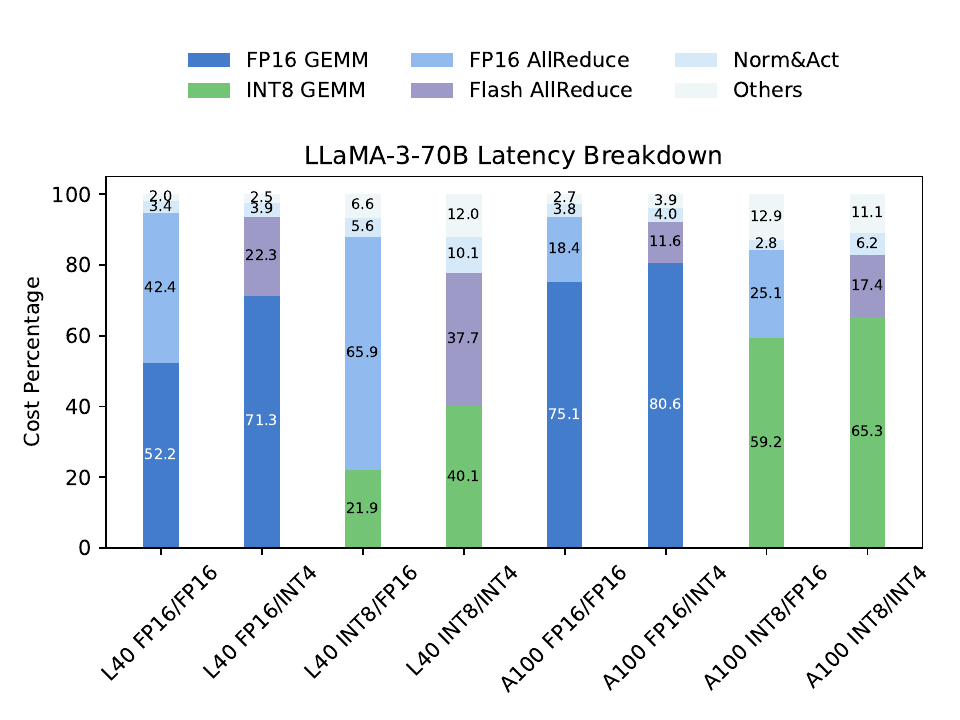}}
\caption{Prefill cost breakdown of LLaMA-3-70B operations with and without Flash Communication, as measured by NSys~\cite{NVIDIANsightSystems}. Tested on 4$\times$L40/A100 GPUs (TP=4) with a batch size of 8, each with 1024 input and 64 output tokens. NCCL~\cite{nvidia2024nccl}'s Ring All-Reduce is applied. The notion of x-ticks (e.g. L40 FP16/FP16) denotes GPU type, model weight precision, and communication precision, respectively.}
\label{fig:comm-overhead}
\end{center}
\vskip -0.2in
\end{figure}

To tackle the scaling difficulties of large language models, the research community now resorts to multiple parallelism strategies across a large group of computing accelerators. Since previous parallelism methods focus on resolving the training challenges, we quickly review these methods for a background check. Particularly, \emph{data parallelism}~\cite{dean2012large,ben2019demystifying} is first introduced to allocate the training samples onto multiple GPUs where each GPU retains a duplicate of the model and processes its own given batch of samples. Synchronization is hence required at the end of each iteration to update the model parameters. In the LLM era, ZeRO~\cite{rajbhandari2020zero} and FSDP~\cite{zhao2023pytorch} renovate data parallelism by sharding models on all devices but virtually rendering a whole model on a single device through All-Gather communication. In contrast, \emph{pipeline parallelism} partitions sequential layers onto different GPUs where point-to-point communication is adopted to transmit activation and gradients. However, it creates data dependency which leads to substantial GPU idle time, called \emph{bubbles}. To improve GPU utilization, GPipe~\cite{huang2019gpipe} schedules microbatches in a pipeline with forward passes and then followed by backward passes. PipeDream~\cite{harlap2018pipedream} proposes one-forward one-backward (1F1B) to further reduce the bubble ratio. Megatron-LM~\cite{narayanan2021efficient} advances PipeDream by allowing each device to perform computation for multiple non-contiguous subsets of layers. Another dimension to split the model is \emph{tensor parallelism} which splits the tensors of each layer and performs All-Reduce to aggregate the activation and gradients from all devices. Megatron-LM~\cite{shoeybi2019megatron,narayanan2021efficient} is such an 
 example that devises delicate conjugate tensor slicing schemes (e.g. column-wise first and row-wise next for the MLP layer) to remove unnecessary synchronization inside a transformer block. 

With the rise of LLMs developed for the long-context scenario, \emph{sequential parallelism}~\cite{korthikanti2023reducing} is proposed to divide activation of LayerNorm and Dropout layers in the sequence dimension as they are sequence-independent. It also jointly combines with tensor-parallelism by replacing two All-Reduce operations with one All-Gather and one Reduce-Scatter to merge the communication cost. However, self-attention and MLP layers are left untouched for sequential parallelism. In this regard, \emph{context parallelism}~\cite{nvidia2024context} is designed to separate all layers in sequence dimension. To break the sequence dependency in self-attention, Ring Attention~\cite{liu2023ring} applies blockwise self-attention and feedforward~\cite{dao2023flashattention2,NEURIPS2023_1bfd87d2} in a distributed environment with point-to-point communication. On top of this, Deepspeed-Ulysses~\cite{jacobs2023deepspeed} exchanges point-to-point communication for All2All for faster speed.

Another emerging direction is sparse architectures represented by mixture-of-experts models~\cite{jiang2024mixtral,qwen_moe,liu2024deepseek}. \emph{Expert parallelism}~\cite{fedus2022switch} parallelizes the experts on different GPUs which requires All2All communication. Deepspeed-MoE~\cite{rajbhandari2022deepspeed} propose hierarchical All2All communication to reduce the number of communication hops.




As large language models continue to scale up, modern frameworks like DeepSpeed~\cite{DeepSpeed}, and Megatron~\cite{MegatronLM}  tend to make joint use of the aforementioned parallelism to accelerate the training process. Nevertheless, they easily meet communication bottlenecks as they require many collective operations. This overhead grows as the model becomes larger.

Meanwhile, the communication bottleneck is also pronounced when serving large language models in the cloud. Constrained by strict service level objectives (SLOs), multiple parallelism schemes are adopted to speed up the inference, where tensor parallelism is the most popular option among all. Besides, due to the lower bandwidth of inference GPUs,  communication can account for more than half of the prefill inference cost where an 80-layer LLaMA-3-70B~\cite{dubey2024llama} carries out 160 all-reduce operations at each forward pass on 4$\times$L40 GPUs, as shown in Figure~\ref{fig:comm-overhead}. Therefore, to enable a faster speed, we must make efficient use of limited intra-node bandwidth.

In this work, we design a novel technique to reduce the communication cost introduced by tensor parallelism without substantially sacrificing accuracy. Our contributions are,

\begin{enumerate}
    \item Through detailed measurements, we unveil the communication bottleneck problem that also recurs in large language model (LLM) inference. For instance, communication can account for up to 65\% of the total latency on NVIDIA L40 GPUs (Fig.~\ref{fig:comm-overhead}).
    \item We design an efficient communication mechanism called \emph{Flash Communication}, which applies low-bit fine-grained quantization on activations to reduce communication volume and employs a \emph{two-step all-reduce} strategy to minimize communication hops.
    \item We implement a fused CUDA kernel called Flash All-Reduce to perform Flash Communication, achieving up to a 2$\times$ reduction in time-to-first-token (TTFT) on NVIDIA L40 GPUs. Even on A100 GPUs with higher communication bandwidth, we observe notable latency reductions, demonstrating the effectiveness of our method.
\end{enumerate}

\section{Related Work}
Before diving into the investigated problem, we cover some fundamental knowledge required for discussion in Appendix~\ref{app:background}. We suggest that readers without prior experience quickly review the content.

Communication efficiency is crucial to distributed training and serving, as it directly affects the total processing time and cost. Several techniques have been proposed to optimize communication in distributed training in recent years, including topology optimization, pipeline optimization, and compression methods. 

\subsection{Topology Optmization}
Topology optimization adjusts communication patterns to match the physical topology of hardware to reduce communication latency/hops, mainly ring-based and tree-based.
Ring-All-Reduce~\cite{baidu_allreduce} organizes workers in a ring topology so that the overall communication latency is constant regardless of the number of workers. Say a worker transmits data of volume $M$ to a group of $N-1$ workers, the total communication volume is $2M(N-1)/N$, which is approximately $2M$ when $N>>1$. However, it doesn't take the physical topology into account, where intra- and inter-node communication have different bandwidths. Hence the average speed depends largely on the lowest bandwidth in such a strategy. Hierarchical Ring-All-Reduce~\cite{jia2018highly} highlights the importance of hierarchical structures in managing overheads, which employs three-phase all-reduce for separate intra- and inter-node communication. Later, 2D-Torus~\cite{mikami2018massively} organizes GPUs in a 2D-grid of $(X,Y)$ so that the inter-node horizontal communication volume is $X$ times smaller than that of hierarchical Ring-All-Reduce.
NCCL~\cite{nvidia2019massively} introduces double binary trees~\cite{sanders2009two} provides logarithmic latency by reducing hops from $2(N-1)$ to $2log(N)$. However, it is more prone to result in suboptimal bandwidth utilization, as only a subset of nodes are engaged in any given communication step. 

With the rise of sparse architectures like mixture-of-experts, All2All collective operation is common for communication in expert parallelism. DeepSpeed-MoE~\cite{rajbhandari2022deepspeed} and HetuMoE~\cite{nie2022hetumoe} both utilize a scalable hierarchical scheme to reduce all-to-all communication hops for faster speed. 

Besides, NVLink SHARP~\cite{nvidia_sharp_release_notes} is a hardware improvement that offloads collective operations from GPUs to the network devices, hence eliminating the need to send data multiple times between endpoints.

\subsection{Pipeline Optimization}  
Pipelining optimization aims to maximize resource utilization with optimized scheduling strategies, mainly by overlapping computation with communication. 
Domino~\cite{wang2024domino} breaks data dependency in Tensor-parallelism by splitting activations row-wisely and weights column-wisely into smaller independent parts.
FLUX~\cite{chang2024flux} divides computation and communication operations into much finer-grained operations and later merges them in a larger kernel to effectively hide communication. DistributedGEMM~\cite{shilabs2024distributedgemm} provides an implementation based on CUTLUSS~\cite{nvidia_cutlass} using P2P communication. 
ScMoE~\cite{cai2024shortcut} implements a shortcut-connected MoE architecture to effectively decouple communication from its conventional sequence, allowing for a substantial overlap.

\subsection{Communication Compression}
Compression techniques like sparsification and quantization are proposed to balance communication reduction with acceptable performance degradation. Sparse Communication~\cite{Aji_2017} observes that gradient updates are mostly close to zero and maps them directly to zero to only exchange sparse matrices among distributed nodes. In contrast, 
DISCO~\cite{qin2024disco} aims to achieve sparse communication by gradually pruning the network to generate sparse features. QSDP~\cite{markov2023quantized} remove FSDP's communication bottleneck by performing both gradient and weight quantization. ZeRO++ \cite{wang2023zero++} applies All-Gather with blockwise weight quantization and an All2All-based gradient quantization to reduce the communication volume when collecting weights and gradients.



\section{Method}

\subsection{Motivation}

\begin{figure}[ht]
\begin{center}
\centerline{\includegraphics[width=\columnwidth]{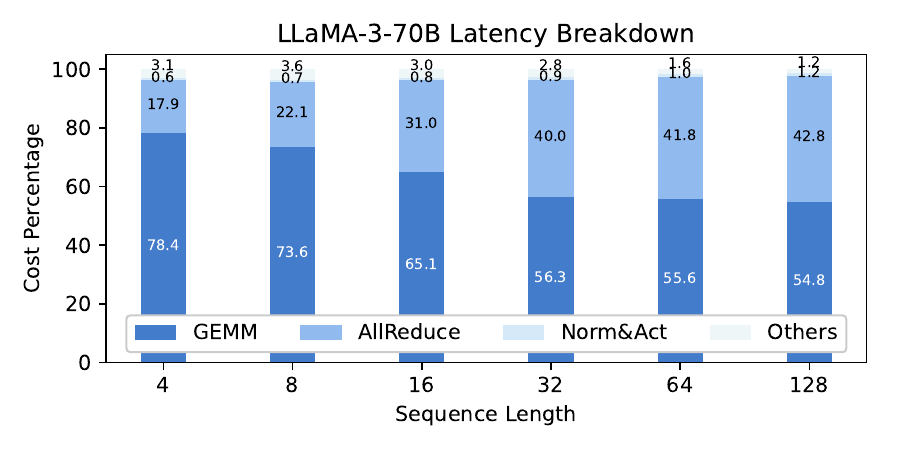}}
\caption{Prefill cost breakdown of LLaMA-3-70B operations at various sequence lengths. Tested on 4$\times$L40 GPUs (TP=4) with a batch size of 8.}
\label{fig:comm-overhead-by-seqlen}
\end{center}
\vskip -0.2in
\end{figure}

Tensor Parallelism (TP) is now supported in almost all mainstream inference frameworks like TensorRT-LLM~\cite{nvidia_tensorrt_llm}, vLLM~\cite{kwon2023efficient}, SGLang~\cite{sglang_project}, and LMDeploy~\cite{2023lmdeploy}, becoming the most adopted scheme in LLM inference. However, TP comes at a non-negligible cost due to heavy communication, which in the case of larger language models creates an excessive communication overhead. For example, the communication overhead of LLaMA-3-70B on L40 GPUs easily meets the bottleneck as the input token length increases, shown by the cost breakdown of LLaMA-3-70B operations in Figure~\ref{fig:comm-overhead-by-seqlen}. Although high-end training-purpose accelerators like NVIDIA A100 where GPUs are connected through NVLink~\cite{nvidia2024nvlink}, the communication overhead still reaches a notable 20\%. We can easily conclude that TP communication is the inference bottleneck. 

For communication optimization, one can think of overlapping the communication with computation to hide overhead but it requires sophisticated design with scheduling which is harder to implement in any given inference framework, for which reason NanoFlow~\cite{zhu2024nanoflow} invents a new serving framework to circumvent the difficulty. On the contrary, compression is a handy option but none of the above-mentioned methods investigated communication quantization for LLM inference. It could be due to the activation quantization challenge posed at the inference stage, which is pointed out by LLM.int8()~\cite{dettmers2022llmint8} and SmoothQuant~\cite{xiao2024smoothquant} that activations are harder to quantize because of outliers. The communication quantization method for training doesn't suffer from this problem as it only quantizes weights and gradients~\cite{wang2023zero++}. Besides, during training, communication quantization degradation can be compensated by further learning. Whereas at inference, the quantization loss is nearly irreversible. 

At the same time, the existing Ring All-Reduce operation adopted in tensor parallelism remains a bottleneck at inference since it is inclined to be constrained by lower bandwidth. Furthermore, to integrate quantization with Ring All-Reduce, also shown by ZeRO++~\cite{wang2023zero++}, it requires $N$ times of sequential quantization and dequantization in a complete Reduce-Scatter, which worsens the latency.

The above issues call for a delicate orchestration of the quantization approach and a better All-Reduce scheme.

\subsection{Flash Communication}

Motivated by the above, we approach the inference challenges in the distributed scenario with quantization and topology optimization. In the paper, we specifically examine tensor parallelism as it is the most popular paradigm. 

\begin{figure}[ht]
\begin{center}
\centerline{\includegraphics[width=\columnwidth]{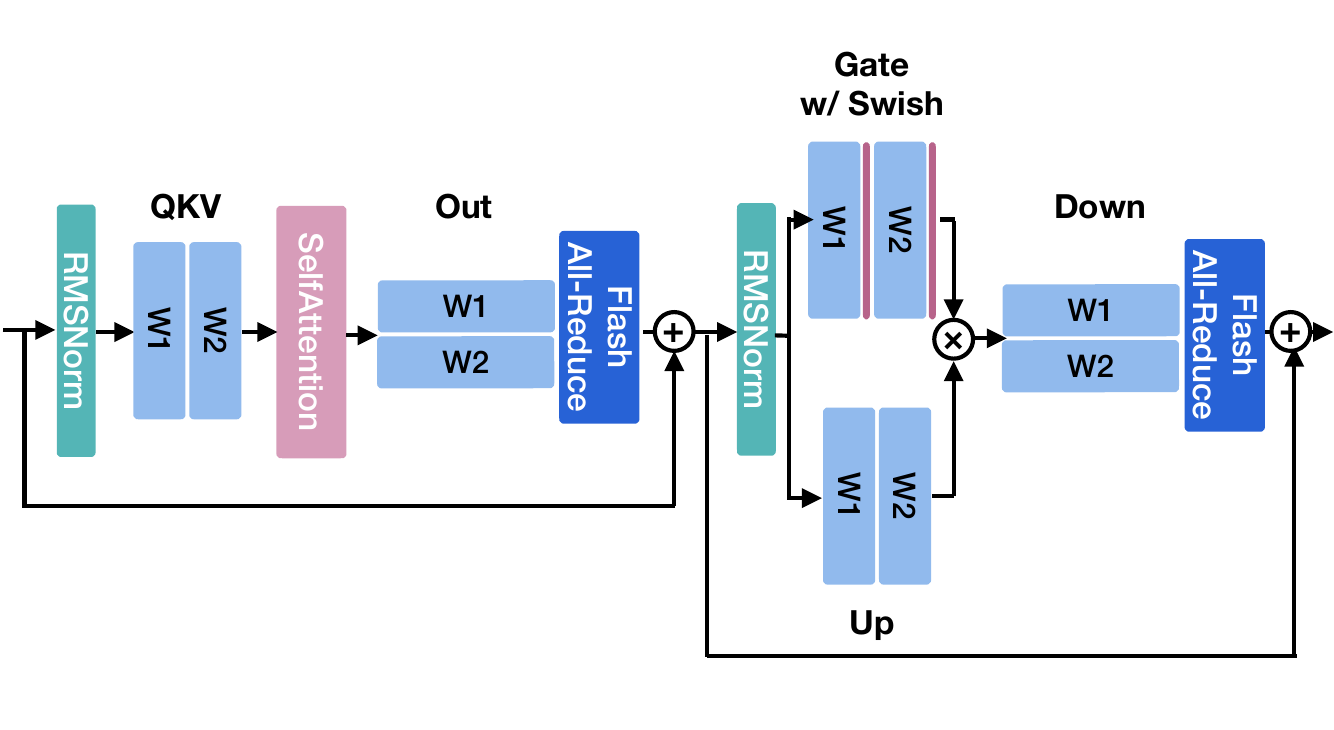}}
\caption{Tensor parallelism for a LLaMA-3 transformer block. Our Flash All-Reduce is applied to speed up communication.}
\label{fig:llama-block-flash-allreduce}
\end{center}
\vskip -0.2in
\end{figure}

Take tensor parallelism in LLaMA-3~\cite{dubey2024llama} as an example, the tensors of QKV projections are first sliced column-wisely and then the output projection row-wisely. After that, an All-Reduce operation is required to collect activations on each device, shown in Fig.~\ref{fig:llama-block-flash-allreduce}. Similarly in the feedforward network, the gate projection and the up projection are split by column and the down projection by row, then another All-Reduce is needed to sum up the activations from both devices. To reduce the communication volume, we are left to compress the activation from the output projection and the down projection.

\subsubsection{Quantization Challenge}

\begin{figure}[ht]
\vskip 0.2in
\begin{center}
\centerline{\includegraphics[width=\columnwidth]{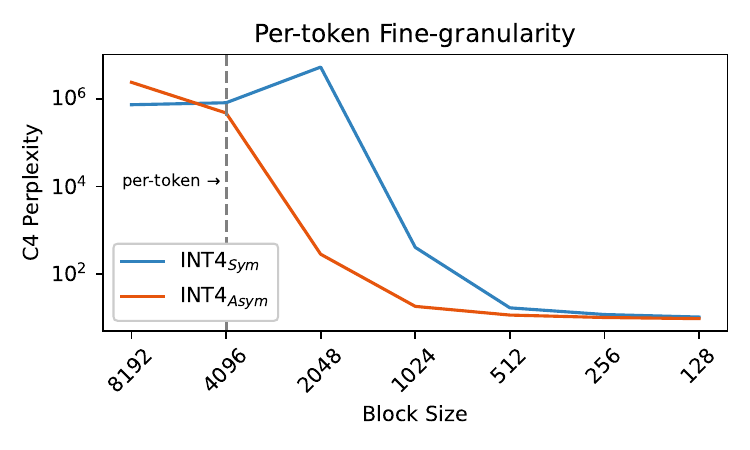}}
\caption{Activation quantization with various block sizes of LLaMA-3-8B on C4. Starting from 4096 (the length of hidden dimension), the granularity becomes finer till 128.}
\label{fig:comm-overhead-long-context}
\end{center}
\vskip -0.2in
\end{figure}

To obtain an optimal trade-off between accuracy and latency, we choose to apply low-bit quantization. From Fig.~\ref{fig:comm-overhead-long-context}, we observe that fine granularity is necessary since per-token quantization at larger block sizes suffers from performance collapse in terms of C4 perplexity, albeit the asymmetric version is relatively better.

\begin{figure}[ht]
\begin{center}
\centerline{\includegraphics[width=\columnwidth]{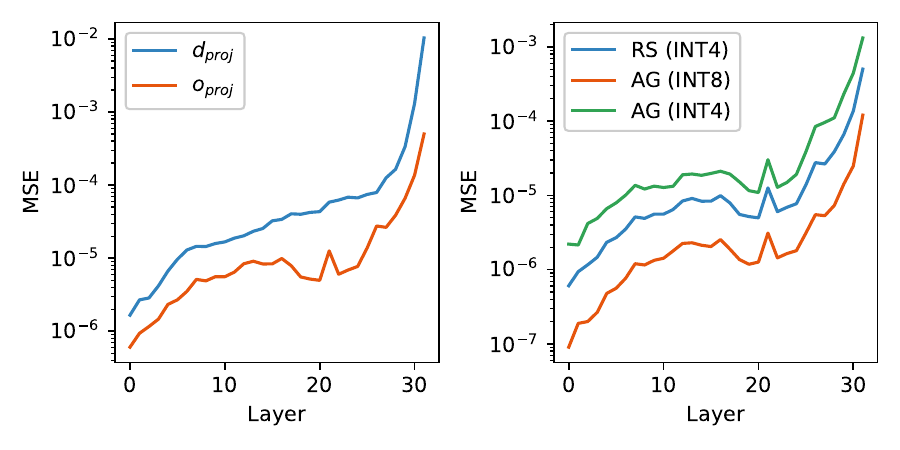}}
\caption{Left: Comparison of $o_{proj}$ and $d_{proj}$ All-Reduce Quantization MSE. Right: MSE of quantization before Reduce-Scatter (RS) vs. All-Gather (AG).}
\label{fig:mse-o-proj-vs-down-proj}
\end{center}
\vskip -0.2in
\end{figure}

However, we discover that it is non-trivial to apply low-bit activation quantization in this scenario. To investigate the quantization sensitivity, we calculate the layerwise mean squared errors (MSE) before and after activation quantization on LLaMA-3-8B, as depicted in Figure~\ref{fig:mse-o-proj-vs-down-proj}. We find that the down projection $d_{proj}$ is much harder to quantize than the output projection $o_{proj}$, as the former MSEs are quite distinct even on a logarithmic scale. This phenomenon is also discovered by ~\cite{li2023fptq,ashkboos2023towards,yu2024super}.

Besides, an All-Reduce comprises a pair of Reduce-Scatter and All-Gather operations, where the quantization corresponding to each operation exhibits different levels of difficulty, see Fig.~\ref{fig:mse-o-proj-vs-down-proj} right. This is as expected since the quantization before Reduce-Scatter only introduces rounding errors while in the case of All-Gather, it includes both rounding and accumulated errors. Alternatively, we could use a higher precision for the quantization before All-Gather to improve accuracy.


\begin{figure*}[ht]
\vskip 0.2in
\begin{center}
\centerline{\includegraphics[width=\textwidth]{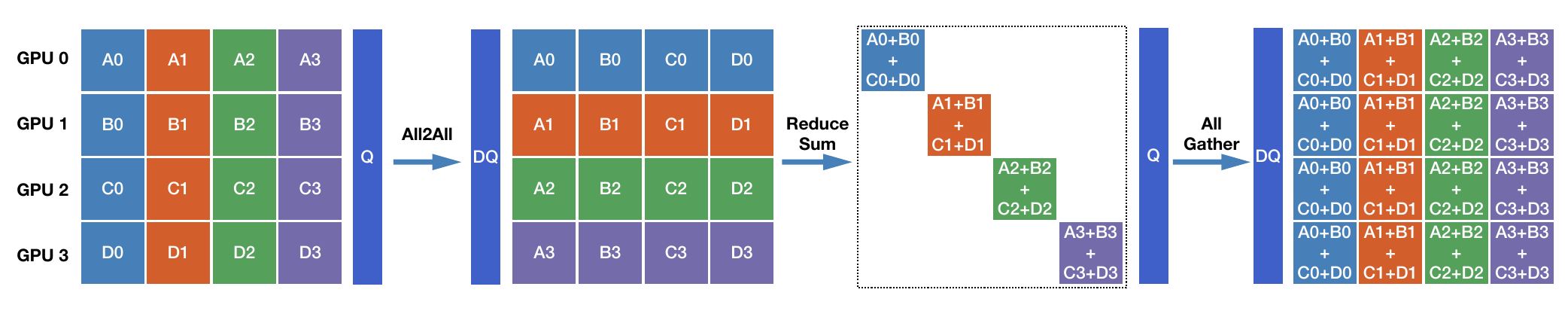}}
\caption{Flash communication's \emph{two-step All-Reduce}. The communication volume is quantized and dequantized only twice.}
\label{fig:flash-comm}
\end{center}
\vskip -0.2in
\end{figure*}

\subsubsection{Algorithm}

Considering the above issues, we design a two-step quantization strategy to replace vanilla Ring All-Reduce, as portrayed in Fig.~\ref{fig:flash-comm}. Its integration with tensor parallelism is shown in Fig.\ref{fig:llama-block-flash-allreduce}.  We name our overall strategy as \emph{two-step All-Reduce}.

Fig.~\ref{fig:flash-comm} illustrates how Flash Communication works. First, we divide the computation volume (activation) on each GPU by the number of ranks. After fine-grained quantization on activation, we perform All2All communication so that each device receives its computation load for reduction. After on-device reduction, the sum is again quantized to speed up the transmission. We then perform All-Gather to collect all results and dequantization to recover float values on each device. This two-step workflow is also formulated by Alg.~\ref{alg:flash-comm}.

\begin{algorithm}[ht]
\caption{Flash All-Reduce}\label{alg:flash-comm}
\begin{algorithmic}
\STATE{\textbf{Input:} Communication volume $M$, world size $N$, chunk size $C$, quantization bit-width $b$, group size $g$}
\STATE{\textbf{Output:} Reduced sum $S^{dq}$}
\STATE{Divide $M$ into $T = \lceil M/C \rceil$ chunks.}
\FOR{$1 \leq i \leq T$}
\STATE{\textcolor{gray}{// Quantize volume to obtain zeros and scales}}
\STATE{$M_i^{q}$, $z_i$, $s_i$ = FinegrainedQuantize($M_i$, $b$, $g$);}
\STATE{\textcolor{gray}{// Each device sends and receives volume from others}}
\STATE{All2All($M_i^{q}$, $z_i$, $s_i$, $N$);}
\FOR{$1 \leq j \leq N$}
\STATE{$M_{i_{j}}^{dq}$ = Dequantize($M_{i_{j}}^{q}$, $z_{i_{j}}$, $_{i_{j}}$);}
\ENDFOR
\STATE{$S_i$ = ReduceSum($M_{i_{0}}^{dq}$, $M_{i_{1}}^{dq}$, $\cdots$, $M_{i_{N}}^{dq}$);}
\STATE{$S_i^{q}$, $z_i^s$, $s_i^s$ = FinegrainedQuantize($S_i,b,g$);} 
\STATE{\textcolor{gray}{// Each device collects the reduced sum from others}}
\STATE {All-Gather($S_i^{q}$, $z_i^s$, $s_i^s$, $N$);}
\FOR{$1 \leq j \leq N$}
\STATE{$S_{i_{j}}^{dq}$ = Dequantize($S_i^{q}$, $z_i^s$, $s_i^s$);}
\ENDFOR
\ENDFOR
\end{algorithmic}
\end{algorithm}

\subsubsection{Kernel Design}
For efficiency, we implement a fused Flash All-Reduce kernel to encompass all the above collective communication operations and quantization steps.
Compared with Ring All-Reduce in Table~\ref{tab:ring-allreduce-vs-flash-allreduce}, Flash All-Reduce cuts quantization-dequantization steps from $N$ to 2, and Reduce/Gather steps from $N-1$ to 1. Although the size of total volumes remains the same, each of our volumes is quantized to lower bits, substantially reducing the amount of data to transmit. We summarize three key aspects in designing our kernel below.

\begin{table}[t]
\setlength{\tabcolsep}{2pt}
\setlength{\extrarowheight}{4pt} 
\caption{Comparison of Ring All-Reduce vs. Flash All-Reduce}
\label{tab:ring-allreduce-vs-flash-allreduce}
\begin{center}
\begin{small}
\begin{sc}
\begin{tabular}{lcc}
\toprule
Method	& Ring All-Reduce	&	Flash All-Reduce	 \\
\midrule
Total Volume	&	$2M(N-1)/N$	&	$2M(N-1)/N$	 \\
Reduce Step	&	$N-1$	& $1$	 \\
Reduce-Scatter	&	$M/N$	&	$M(N-1)/N$\\
Gather Step & $N-1$ &  $1$ \\
All-Gather       &	$M/N$	& $M(N-1)/N$ \\
QDQ Step & $N$ & $2$ \\
\bottomrule
\end{tabular}
\end{sc}
\end{small}
\end{center}
\end{table}

\begin{figure}[ht]
\begin{center}
\centerline{\includegraphics[width=\columnwidth]{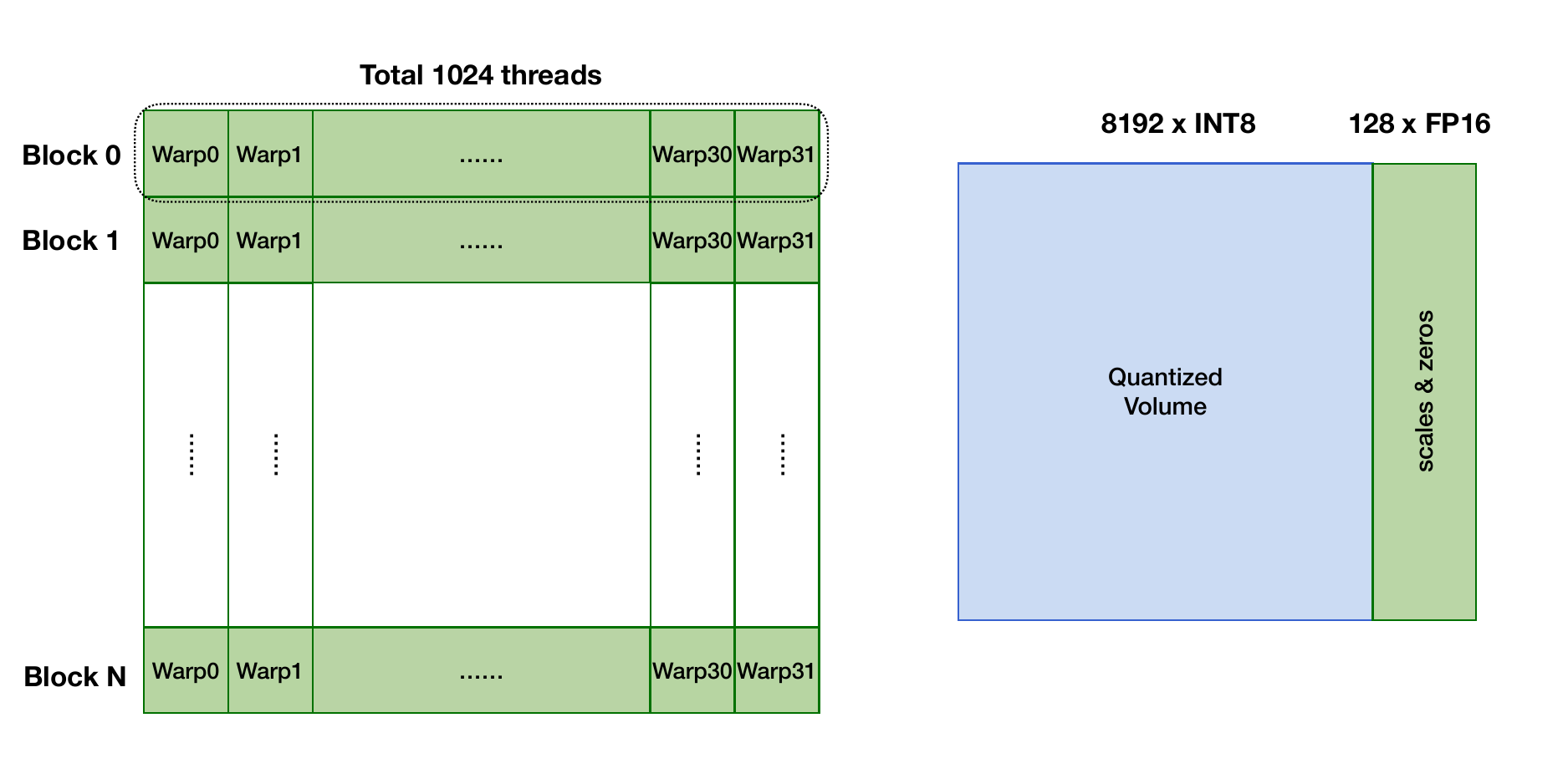}}
\caption{Thread mapping of fast fine-grained quantization.}
\label{fig:warp-thread}
\end{center}
\vskip -0.2in
\end{figure}

\textbf{Fast Fine-grained Quantization}
The total communication volume $M$ for each rank is divided into $T$ chunks for transmission. Given a chunk size $C$, we draw how GPU threads are organized in parallel to process the chunk information in Fig.~\ref{fig:warp-thread}. A chunk is split into $N$ blocks and each block corresponds to 32 warps, where each warp is a collection of 32 threads and each thread can process eight FP16 elements. Take our asymmetric quantization with a group size of 128 as an example, we perform quantization on each group of 128 elements using 16 threads. Specifically, we leverage the CUDA API function \texttt{\_\_shfl\_xor\_sync} to iteratively exchange information among these warp threads to achieve max/min reduction efficiently.

\textbf{Fast Communication.} Instead of calling All2All primitive, we utilize GPU peer direct memory access from CUDA Runtime API~\cite{nvidia_cuda_peer_device_memory_access} to transmit quantized volume, where we can directly fetch data from different ranks, substantially boosting the communication speed.

\begin{figure}[ht]
\begin{center}
\centerline{\includegraphics[width=\columnwidth]{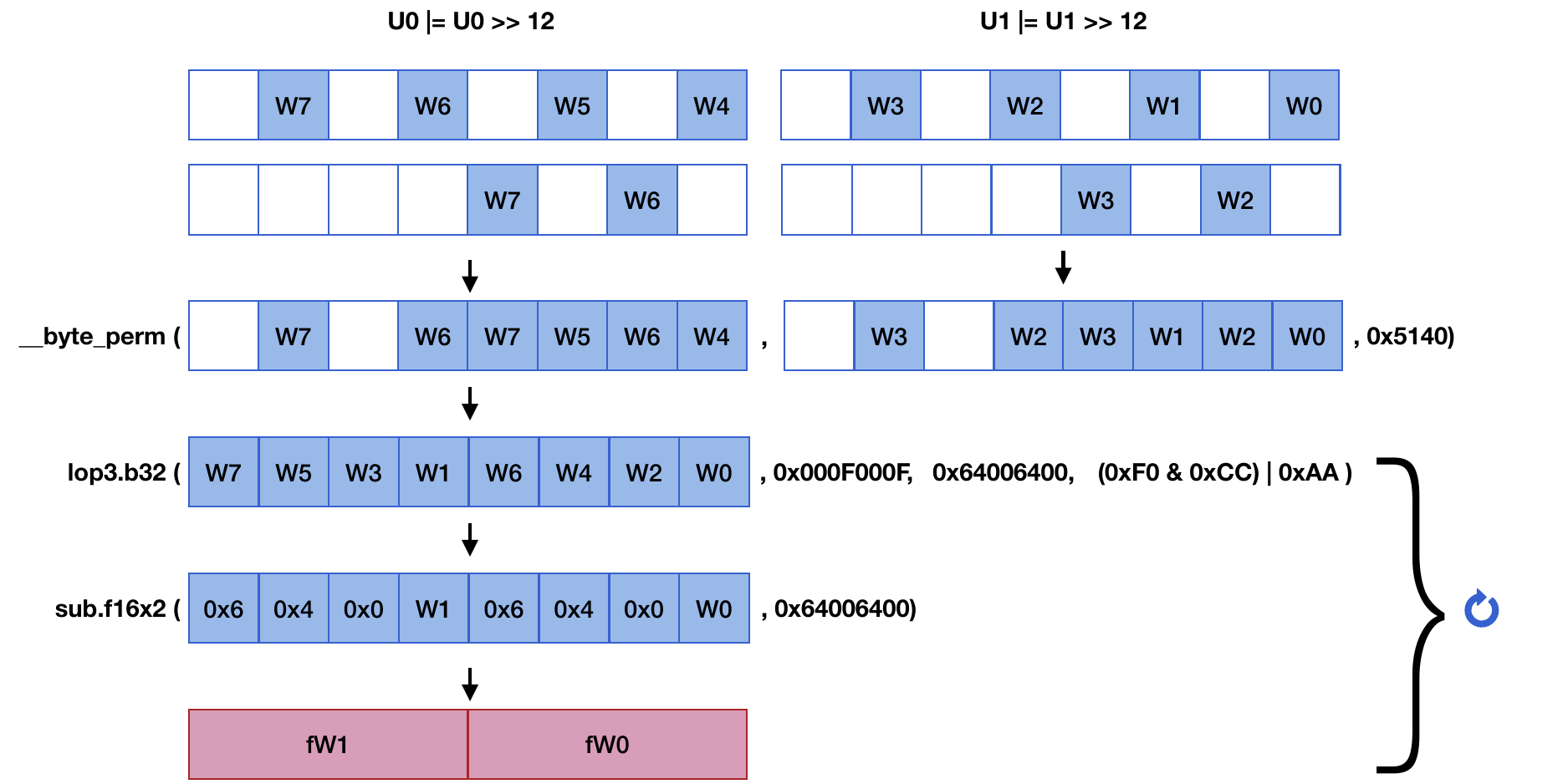}}
\caption{Fast INT4 packing and dequantization. For simplicity, only the dequantization of W1 and W0 is shown. An empty square means zero.}
\label{fig:fast-pack-int4-dequant}
\end{center}
\vskip -0.2in
\end{figure}

\textbf{Fast Dequantization.} Once quantization volumes are received, we have to dequantize them into FP16 for sum reduction. As na\"i{ve} INT4 to FP16 conversion incurs overhead, we utilize the dequantization layout in ~\cite{kim2022says}. To coordinate its ordering online, we also employ fast INT4 packing~\cite{2023lmdeploy}, as illustrated in Fig.~\ref{fig:fast-pack-int4-dequant}. Given that two 32-bit unsigned integers \texttt{U0} and \texttt{U1} holding 4 INT4-quantized activations (each stored in the lower 4 bits out of 8 bits) to transmit, we first perform the right shift by 12 bits, and then apply bitwise OR to itself. Later we select the target bits from these two integers with CUDA Math API \texttt{\_\_byte\_perm}. In this way, we can pack 8 4-bit integers in a convenient order to dequantize. Next we apply   \texttt{lop3.b32}\footnote{\url{https://docs.nvidia.com/cuda/parallel-thread-execution/\#logic-and-shift-instructions-lop3}} to perform logical operation \texttt{(0xF0 \& 0xCC)|0xAA} on the packed variable with mask \texttt{0x000F000F} and \texttt{0x64006400}, then we subtract it with  \texttt{0x64006400},  which effectively represents \texttt{W1} and \texttt{W0} in FP16. The dequantization can be performed iteratively by varying the masks for the rest INT4 integers.

\textbf{INT6 Quantization.}
Given that quantization prior to All-Gather leads to greater loss, as illustrated in Figure~\ref{fig:mse-o-proj-vs-down-proj} (left), we opt for an INT8 bit-width for All-Gather operations and maintain INT4 for ReduceSum, effectively creating an INT6 solution. As later shown in Table~\ref{tab:mmlu-c4-wiki-comm-prec}, the INT6 configuration strikes a commendable balance between performance and communication efficiency.

\section{Experiments}

\subsection{Setup}

Unless otherwise noted,  we use an input token length of 1024 and an output token length of 64 for the inference measurement. Latencies are tested on NVIDIA L40 and A100 SXM GPUs. The baseline uses FP16 for communication. 

\subsection{Accuracy Comparison}

\textbf{FP16 Weights.} We evaluate the accuracy of LLaMA-2 and LLaMA-3 models on PIQA~\cite{bisk2020piqa}, ARC\textsubscript{C} and ARC\textsubscript{E}~\cite{clark2018think}, HellaSwag \cite{zellers2019hellaswag}, WinoGrande \cite{sakaguchi2021winogrande} in various communication quantization bit widths, shown in Table~\ref{tab:mmlu-comm-quant}. In all cases, asymmetric INT8 quantization obtains the best accuracies. Asymmetric INT4 is also better than symmetric INT4. C4~\cite{C4} and WikiText~\cite{wikitext103} results are shown in Table~\ref{tab:c4-wiki-fp16} of Appendix~\ref{app:c4-wiki-exp}.

\begin{table}[ht]
\setlength{\tabcolsep}{2pt}
\caption{Accuracy of LLaMA models with various communication quantization strategies. All model weights are kept in FP16 precision. Prec: Communication precision. All INT quantization is asymmetrical. HS: HellaSwag, WG: WinoGrande.}
\label{tab:mmlu-comm-quant}
\vskip 0.15in
\begin{center}
\begin{small}
\begin{sc}
\begin{tabular}{l*{7}{c}}
\toprule
Model	&	Prec	&	PIQA	&	ARC$_C$	&	ARC$_E$	&	HS	&	WG	&	Avg \\
\midrule
2-7B	&	FP16	&	79.11	&	46.33	&	74.58	&	76.01	&	69.30	&	69.07  \\
	&	INT8	&	79.11	&	45.99	&	74.75	&	76.10	&	69.06	&	69.00  \\
	&	INT6	&	78.78	&	45.99	&	74.79	&	75.75	&	68.75	&	68.81  \\
	&	INT4	&	78.02	&	45.82	&	74.49	&	75.63	&	67.25	&	68.24  \\
    \midrule
2-13B	&	FP16	&	80.52	&	49.15	&	77.48	&	79.39	&	72.14	&	71.74  \\
	&	INT8	&	80.69	&	49.06	&	77.61	&	79.34	&	71.59	&	71.66  \\
	&	INT6	&	79.98	&	49.32	&	77.06	&	79.17	&	71.11	&	71.33  \\
	&	INT4	&	79.60	&	48.21	&	76.89	&	78.92	&	71.90	&	71.10  \\
    \midrule
2-70B	&	FP16	&	82.70	&	57.34	&	81.02	&	83.8	&	77.98	&	76.57  \\
	&	INT8	&	82.75	&	57.68	&	80.93	&	83.8	&	77.98	&	76.63  \\
	&	INT6	&	82.48	&	57.00	&	80.85	&	83.77	&	76.87	&	76.19  \\
	&	INT4	&	82.92	&	57.25	&	80.43	&	83.6	&	77.27	&	76.29  \\
    \midrule
3-8B	&	FP16	&	80.79	&	53.41	&	77.69	&	79.16	&	72.77	&	72.76  \\
	&	INT8	&	80.58	&	52.47	&	77.40	&	79.09	&	73.09	&	72.53  \\
	&	INT6	&	79.98	&	51.37	&	77.65	&	78.73	&	73.16	&	72.18  \\
	&	INT4	&	80.09	&	51.02	&	75.84	&	78.11	&	70.48	&	71.11  \\
    \midrule
3-70B	&	FP16	&	84.55	&	64.33	&	85.86	&	84.89	&	80.35	&	80.00  \\
	&	INT8	&	84.55	&	63.91	&	85.82	&	84.9	&	80.82	&	80.00  \\
	&	INT6	&	84.11	&	61.69	&	85.44	&	84.87	&	80.35	&	79.29  \\
	&	INT4	&	83.13	&	61.35	&	83.33	&	84.69	&	78.93	&	78.29  \\
\bottomrule
\end{tabular}
\end{sc}
\end{small}
\end{center}
\end{table}

\begin{figure*}[ht]
  \centering
  \subfigure{
  \includegraphics[width=0.45\linewidth]{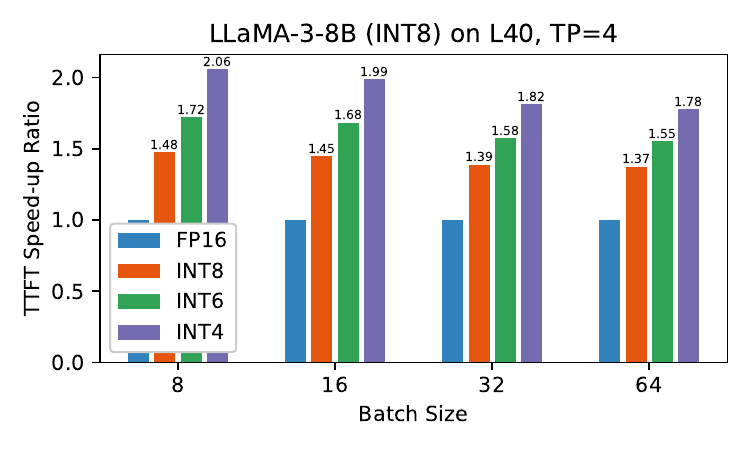}
  }  
  \subfigure{
  \includegraphics[width=0.45\linewidth]{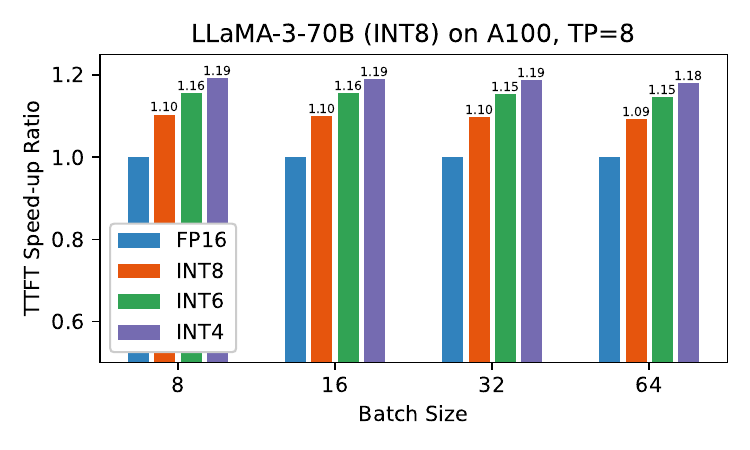}
  }
   \caption{TTFT speed-up ratio of 8-bit LLaMA-3-8B under various communication quantization bit widths on L40 with TP=4 (left), and 8-bit LLaMA-3-70B on A100 with TP=8 (right). }
  \label{fig:ttft-e2e-comparison-l40-and-a100}
\end{figure*}

\textbf{INT8 Weights.} As model weights are quantized, the impact of communication overhead becomes more pronounced. This observation motivates us to explore communication quantization in this context. We begin by quantizing the weights of the LLaMA model using Smoothquant~\cite{xiao2024smoothquant}. Subsequently, we implement fine-grained communication quantization to assess its impact on performance as detailed in Table~\ref{tab:mmlu-c4-wiki-comm-prec}. Results on C4~\cite{C4}, WikiText-2~\cite{wikitext103} is shown in  Appendix~\ref{app:c4-wiki-exp}.

\begin{table}[ht]
\setlength{\tabcolsep}{2pt}
\caption{Accuracy of LLaMA models with various communication quantization strategies (denoted as `Comm Prec'). All model weights are quantized into INT8 precision with SmoothQuant ($\alpha=0.85$ except $0.9$ for LLaMA2-70B). Prec: Communication precision. All INT quantization is asymmetrical. }
\label{tab:mmlu-c4-wiki-comm-prec}
\begin{center}
\begin{small}
\begin{sc}
\begin{tabular}{l*{7}{c}*{2}{H}}
\toprule
Model	&  Prec	&	PIQA	&	ARC$_{C}$	&	ARC$_{E}$	&	HS	&	WG	&	Avg	&	C4	&	WikiText2\\
\midrule
2-7B	&	FP16	&	79.00	&	46.16	&	74.24	&	75.89	&	68.75	&	68.81	&	7	&	5.5\\
	&	INT8	&	79.27	&	45.65	&	74.37	&	75.89	&	68.51	&	68.74	&	7.00	&	5.50\\
	&	INT6	&	78.56	&	45.39	&	74.75	&	75.75	&	68.67	&	68.62	&	7.10	&	5.59\\
	&	INT4	&	77.53	&	45.31	&	74.20	&	75.51	&	68.59	&	68.23	&	7.24	&	5.69\\
    \midrule
2-13B	&	FP16	&	80.25	&	49.49	&	77.27	&	79.37	&	71.59	&	71.59	&	6.51	&	4.92\\
	&	INT8	&	80.41	&	49.32	&	77.53	&	79.21	&	72.22	&	71.74	&	6.51	&	4.92\\
	&	INT6	&	80.09	&	49.40	&	77.02	&	79.14	&	71.74	&	71.48	&	6.55	&	4.97\\
	&	INT4	&	79.11	&	49.06	&	76.30	&	78.89	&	71.27	&	70.93	&	6.63	&	5.03\\
    \midrule
2-70B	&	FP16	&	83.08	&	57.94	&	80.98	&	83.54	&	77.66	&	76.64	&	5.54	&	3.35\\
	&	INT8	&	83.08	&	57.76	&	80.68	&	83.55	&	78.14	&	76.64	&	5.54	&	3.35\\
	&	INT6	&	82.81	&	57.94	&	80.85	&	83.78	&	76.80	&	76.44	&	5.57	&	3.39\\
	&	INT4	&82.92	&	56.83	&	80.35	&	83.41	&	78.14	&	76.33	&	5.62	&	3.43	\\
    \midrule
3-8B	&	FP16	&	80.36	&	51.96	&	77.69	&	78.71	&	73.48	&	72.44	&	9	&	6.24\\
	&	INT8	&	80.09	&	52.90	&	77.57	&	78.79	&	73.09	&	72.49	&	9.01	&	6.25\\
	&	INT6	&	80.03	&	52.56	&	79.12	&	78.38	&	71.82	&	72.38	&	9.33	&	6.49\\
	&	INT4	&	78.94	&	50.17	&	77.10	&	77.87	&	70.56	&	70.93	&	9.85	&	6.84\\
    \midrule
3-70B	&	FP16	&	84.44	&	63.99	&	85.56	&	84.55	&	79.72	&	79.65	&	6.82	&	2.96\\
	&	INT8	&	83.84	&	63.48	&	85.56	&	84.67	&	79.79	&	79.47	&	6.82	&	2.96\\
	&	INT6	&	83.57	&	61.26	&	83.67	&	84.66	&	80.58	&	78.75	&	6.95	&	3.13\\
	&	INT4	&	82.70	&	61.60	&	83.29	&	84.51	&	77.82	&	77.98	&	7.13	&	3.32	 \\
\bottomrule
\end{tabular}
\end{sc}
\end{small}
\end{center}
\vskip -0.15in
\end{table}

\subsection{Latency and Throughput Performance}

Fig.~\ref{fig:ttft-e2e-comparison-l40-and-a100} illustrates weight-quantized LLaMA-3-8B and LLaMA-3-70B's TTFT comparison with and without Flash communication. The lowest quantization bit yields the most gain, i.e. \textbf{2.06$\times$} and \textbf{1.19$\times$} on L40 and A100 respectively. More measurements are listed in Appendix~\ref{app:latency}.

\section{Ablation Study}

\subsection{Integer vs. Low-bit Float}
Table.~\ref{tab:c4-fp-vs-int} shows the difference between INTx and FPx communication quantization. In general, INT8 performs comparably with FP8, while the asymmetric version of INT8 is the best among all. FP6 is a mixed version of FP8~\cite{micikevicius2022fp8} and FP4~\cite{rouhani2023microscaling}, which is a fair comparison with similarly mixed INT6.

\begin{table}[ht]
\caption{LLaMA models' C4 Perplexity with INTx vs FPx quantization with a group size of 128. Weights are in FP16.}
\label{tab:c4-fp-vs-int}
\begin{center}
\begin{small}
\begin{sc}
\begin{tabular}{l*{5}{c}}
\toprule
Prec	&	2-7B	&	2-13B	&	2-70B	&	3-8B	&	3-70B \\
\midrule
FP8$_{Sym}$	&	6.98	&	6.47	&	5.52	&	8.90	&	6.75 \\
INT8$_{Sym}$	&	6.98	&	6.47	&	5.52	&	8.90	&	6.74 \\
INT8$_{Asym}$	&	\textbf{6.98}	&	\textbf{6.47}	&	\textbf{5.52}	&	\textbf{8.89}	&	\textbf{6.74} \\
\midrule
FP6$_{Sym}$	&	7.09	&	6.52	&	5.56	&	9.22	&	6.87 \\
INT6$_{Sym}$	&	7.15	&	6.57	&	5.59	&	9.48	&	6.94 \\
INT6$_{Asym}$	&	\textbf{7.08}	&	\textbf{6.51}	&	\textbf{5.55}	&	\textbf{9.20}	&	\textbf{6.86} \\
\midrule
FP4$_{Sym}$	&	7.24	&	6.60	&	5.61	&	9.72	&	7.07 \\
INT4$_{Asym}$ &	7.50	&	6.71	&	5.69	&	10.51	&	7.30 \\
INT4$_{Asym}$	&	\textbf{7.21}	&	\textbf{6.58}	&	\textbf{5.60}	&	\textbf{9.68}	&	\textbf{7.04} \\
\bottomrule
\end{tabular}
\end{sc}
\end{small}
\end{center}
\vskip -0.15in
\end{table}

\subsection{Na\"{i}ve vs. Rotation-based Quantization}
As previously shown in Fig.~\ref{fig:comm-overhead-long-context}, the C4 perplexity of coarse quantization suffers performance collapse, while fine-grained quantization gradually resolves the problem as the group size increases. We investigate whether the Hadamard transform popularized by QuaRot~\cite{ashkboos2024quarot} could alleviate the issue in the coarse setting in Table~\ref{tab:fg-vs-coarse-and-vallina-vs-quarot}. It turns out that Hadamard transform with coarse quantization readily performs well, however, it loses its advantage in finer granularity cases as compared with na\"{i}ve asymmetric quantization. Besides, it doesn't exhibit gains on FP8. 



\begin{table}[ht]
\setlength{\tabcolsep}{2pt}
\caption{LLaMA models' C4 perplexity with communication quantization (na\"ive vs. rotation) at various granularity levels and precision formats. Asymmetric quantization is used for INT4 and symmetric for FP8. HT: Hadamard Transform}
\label{tab:fg-vs-coarse-and-vallina-vs-quarot}
\begin{center}
\begin{small}
\begin{sc}
\begin{tabular}{lcH*{2}{c}|*{2}{c}}
\toprule
Models	&	Group 	&	C4	&	INT4	&	w/ HT &	FP8	&	w/ HT \\
\midrule
3-8B	&	8192	&	PPL	&	2363367.8	&	10.61	&	8.91	&	8.96\\
	&	2048	&	PPL	&	284.56	&	10.11	&	8.91	&	8.96\\
	&	128	&	PPL	&	9.68	&	\textbf{9.67}	&	\textbf{8.90}	&	8.94\\
\midrule
3-70B	&	8192	&	PPL	&	7417.79	&	7.86	&	6.75	&	6.81\\
	&	1024	&	PPL	&	10.47	&	7.76	&	6.75	&	6.81\\
	&	128	&	PPL	&	\textbf{7.04}	&	7.64	&	\textbf{6.75}	&	6.82\\
\midrule
2-7B	&	8192	&	PPL	&	47601520	&	7.86	&	7.01	&	7.01\\
	&	1024	&	PPL	&	8.78	&	7.35	&	6.98	&	7.01\\
	&	128	&	PPL	&	\textbf{7.21}	&	7.24	&	\textbf{6.98}	&	7.01\\
\midrule
2-13B	&	8192	&	PPL	&	306.39	&	6.80	&	6.47	&	6.49\\
	&	1024	&	PPL	&	7.43	&	6.68	&	6.47	&	6.49\\
	&	128	&	PPL	&	\textbf{6.58}	&	6.62	&	\textbf{6.47}	&	6.49\\
\midrule
2-70B	&	8192	&	PPL	&	49.96	&	5.71	&	5.52	&	5.54\\
	&	1024	&	PPL	&	6.17	&	5.67	&	5.52	&	5.54\\
	&	128	&	PPL	&	\textbf{5.60}	&	5.64	&	\textbf{5.52}	&	5.54 \\
\bottomrule
\end{tabular}
\end{sc}
\end{small}
\end{center}
\end{table}

\subsection{Flash All-Reduce vs. Ring All-Reduce}

\begin{figure}[ht]
\begin{center}
\centerline{\includegraphics[width=\columnwidth]{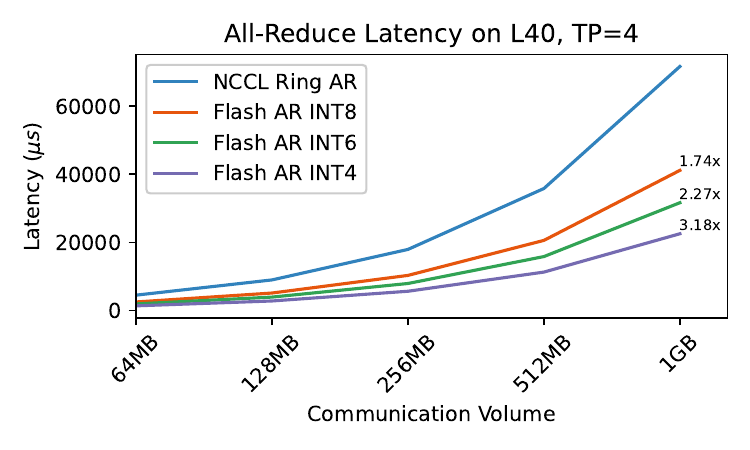}}
\caption{Flash Communication's All-Reduce (Flash AR) Performance compared with NCCL's ring version. NCCL's latency is tested with \texttt{nccl-test}~\cite{NVIDIA_nccl_tests}.}
\label{fig:all-reduce-perf}
\end{center}
\end{figure}

Assembling several boosting techniques, the speed of our Flash All-Reduce kernel surpasses that of Ring All-Reduce by a large margin. Fig.~\ref{fig:all-reduce-perf} exhibits the latency measurement given a certain amount of communication volume. When the communication volume becomes obvious (e.g. larger than 64MB), our quantized All-Reduce is crucial to reduce the cost, where the INT4 version brings at most \textbf{3.18$\times$} kernel latency reduction. Noticeably, the mixed precision version INT6 obtains a good trade-off between INT8 and INT4.

\begin{figure}[ht]
\begin{center}
\centerline{\includegraphics[width=\columnwidth]{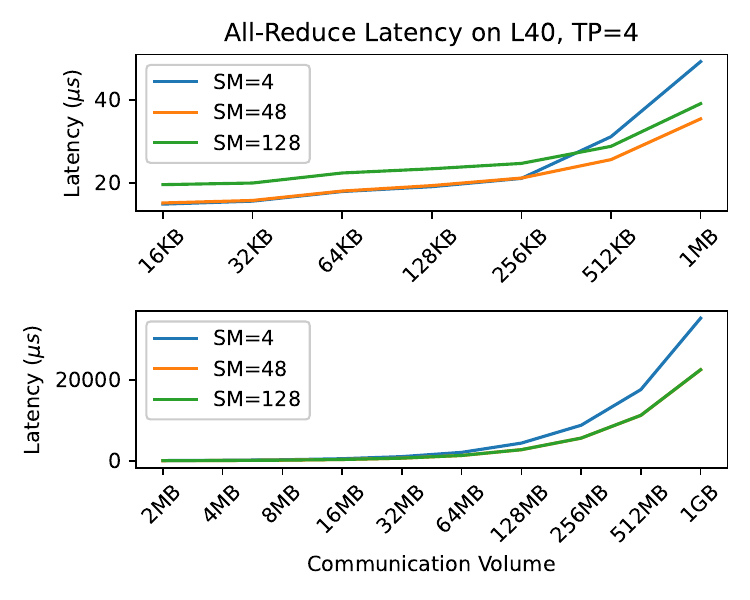}}
\caption{The number of SMs affects the communication latency at different sizes of communication volume. SM=48 is similar to SM=128 in larger volumes.}
\label{fig:all-reduce-perf-sm}
\end{center}
\end{figure}

We further show that the number of streaming processors (SMs) matters in Fig.~\ref{fig:all-reduce-perf-sm}. When the communication volume is of small size, a smaller number of SMs is beneficial as less kernel launch and inter-block synchronization overhead is produced. When the volume gets larger, more SMs are required for calculation. A configuration of 48 SMs strikes a better balance between communication and computation.

\section{Conclusion}

Our work presents a novel technique to reduce the communication volume associated with tensor parallelism while maintaining accuracy. Our key contributions include a comprehensive analysis that reveals the communication bottleneck in Large Language Model (LLM) inference, the design of a fast communication mechanism known as Flash Communication, and the demonstration of its implementation, which has been shown to achieve up to a 2$\times$ TTFT reduction. Flash Communication employs fine-grained quantization on activations and a two-step All-Reduce strategy to decrease communication volumes significantly. We have conducted extensive experiments on NVIDIA L40 and A100 GPUs across various configurations and with several state-of-the-art LLMs, which have consistently demonstrated the effectiveness of our approach. These findings address a critical challenge in parallel computing and pave the way for more efficient and scalable LLM inference.

\clearpage
\newpage
\bibliography{flashcomm}
\bibliographystyle{icml2024}

\newpage
\appendix
\onecolumn

\section{Background}~\label{app:background}

\subsection{GPU Topology}

\begin{figure}[ht]
\begin{center}
\centerline{\includegraphics[width=0.5\columnwidth]{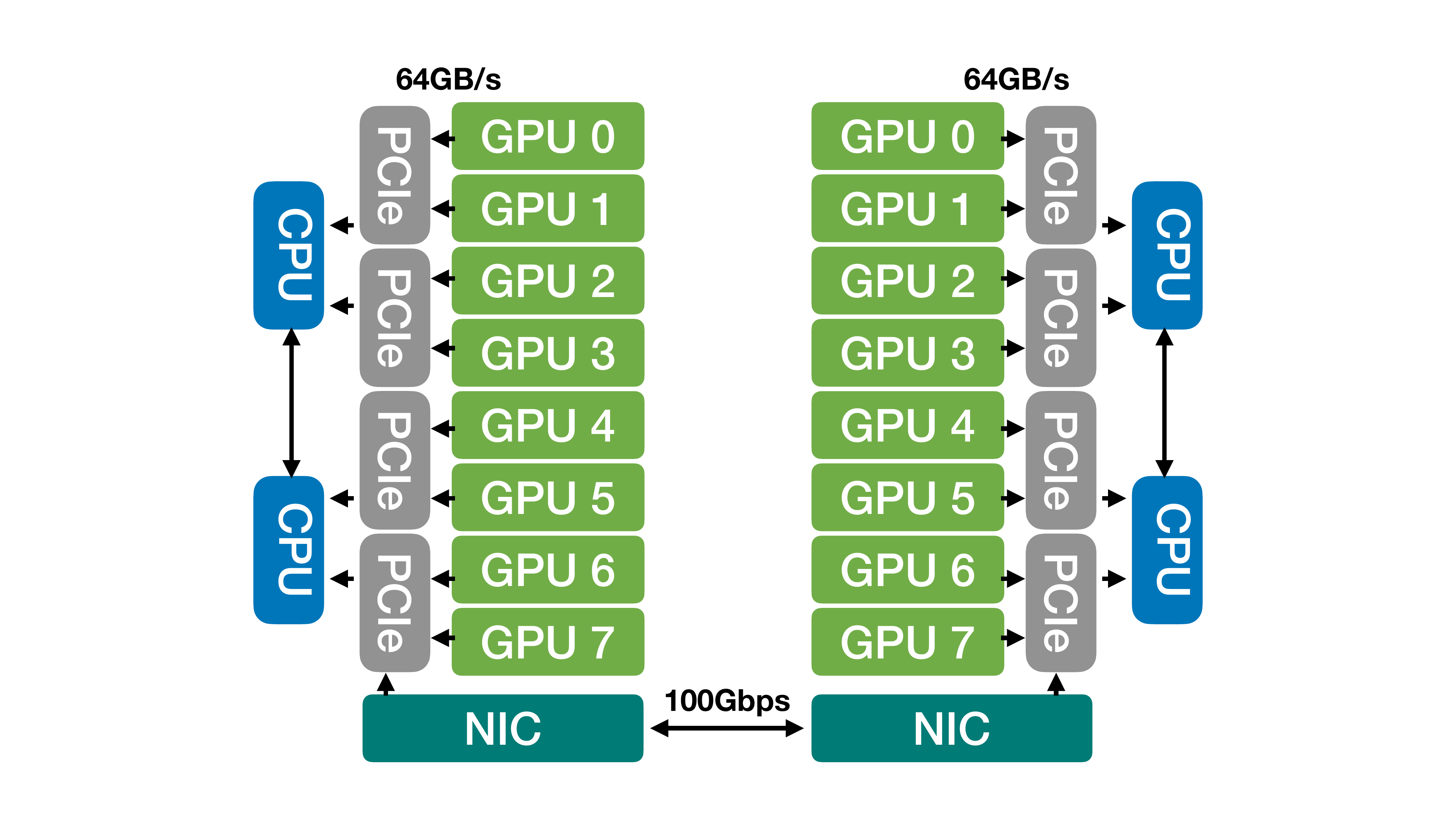}}
\caption{Physical topology of two NVIDIA  8$\times$ L40 GPU nodes connected for inference. Each node has 8 GPUs interconnected with PCI Switches and 2 NUMA nodes. For simplicity, NIC is shown to only connect with the last PCI.}
\label{fig:l40-topo}
\end{center}
\vskip -0.2in
\end{figure}

Modern inference GPUs are connected via various hardware configurations. Figure~\ref{fig:l40-topo} shows a typical simplified physical topology where every node contains 8 GPUs. Every two adjacent GPUs are connected via a PCIe of 64GB/s bandwidth~\cite{nvidia2024l40}. Cross-GPU communication may take several different paths, e.g.  GPU 0 to 1  has the shortest route, but from GPU 0 to 4 it has to go across two NUMA (Non-uniform memory access) nodes. Cross-node communication relies on NICs (Network interface cards) that transmit data via Ethernet, whose bandwidth is usually 100Gbps.

For high-performance large-scale training, high-end GPUs like A100 SXM GPUs enjoy a much wider bandwidth due to the combination use of NVLink~\cite{nvidia2024nvlink}, NVSwitch, and InfiniBand NIC. Each GPU in the same node directly connected with all other GPUs could reach 600GB/s via NVSwitch, and inter-node communication could have a bandwidth of 200Gbps. These advanced hardware configurations tremendously accelerate the training of large language models. 

Warp scheduling is critical for GPU utilization. L40 is shipped with 142 streaming multiprocessors (SM) while A100 has 108. A warp consists of a group of 32 threads, which is the minimum scheduling unit for SM. Multiple warps can be simultaneously executed on an SM.

\subsection{Collective Communication}
\textbf{Libraries.} Due to the hierarchical design of GPU clusters, collective communication methods are crucial in distributed training and inference. To synchronize the workloads across GPUs, communication libraries like NCCL~\cite{nvidia2024nccl}, MSCCL~\cite{microsoft2024msccl}, HiCCL~\cite{hidayetoglu2024hiccl}, Gloo~\cite{gloo}, and Horovod~\cite{sergeev2018horovod} are developed to provide efficient collective communication operations for a group of devices. These libraries usually hide the physical topology and organize GPUs in a ring~\cite{mikami2018massively,jia2018highly,ying2018image} or a tree. In ring-based topology, GPUs are connected hand by hand to create a logical circle, which maximizes the utilization of the full bandwidth. In contrast, tree-based topology, especially double binary tree~\cite{sanders2009two}, guarantees logarithmic communication hops. Therefore, it is more beneficial to use ring-based communication for intra-GPUs and a tree-based approach for inter-GPU clusters. 

\textbf{Operations.}  Collective operations such as broadcast, aggregation (Reduce/All-Reduce/Reduce-Scatter), collection (Gather/All-Gather), and All2All are shipped out-of-box in most collective communication libraries. For instance, NCCL provides a series of such collective operations~\cite{nvidia2024collective} where each rank processes or transmits the same amount of data. Reduce-Scatter sums data across nodes and then scatters the result to corresponding nodes. All-Gather collects data from all nodes to all nodes. All-Reduce is a many-to-many reduce operation, where the same reduce operation is applied on all nodes. All2All exchanges data between all nodes, with each node sending and receiving an equal amount of data. All2All can be implemented with multiple point-to-point communication operations. 

\subsection{Quantization Fundamentals}~\label{app:quant}

Quantization is a mapping from floating numbers to integers. We utilize asymmetric quantization which is formulated below,
\begin{equation}
   s = \frac{X_{max} - X_{min}}{2^n -1 },  z = \ceil{\frac{-X_{min}}{s}}
\end{equation}
\begin{equation}
   Q(X) = clamp(\ceil{X / s } + z, 0, 2^n - 1)
\end{equation}
where $X_{max}$ and $X_{min}$ denotes the maximum and minimum value of $X$, $n$ is the quantization bit-width, $s$ is called the scale and $z$ the zero point. $Q(x)$ quantizes float $X$ to integer to the target bitwidth.

Symmetric quantization is formulated as follows,
\begin{equation}
   s = \frac{\vert X \vert_{max} }{2^{n -1 } -1}
\end{equation}
\begin{equation}
   Q(X) = clamp(\ceil{X/s}, -2^{n-1}, 2^{n-1} - 1)
\end{equation}

\textbf{IEEE 754 standards for FP16.} IEEE 754~\cite{ieee754} FP16 includes 16 bits in total, which comprises 1 bit for the sign (S), 5 bits for the exponent (E), and 10 bits for the mantissa or fraction (F). The bias for the exponent is 15, which means that the actual exponent value must be added to 15 to get the stored exponent value. Also, notice there's an assumed leading 1 in the fractional part.

\textbf{FP8 and FP4 Format.} FP8~\cite{micikevicius2022fp8} format is designed to advance FP16 with two encodings, E4M3 (4-bit exponent and 3-bit mantissa) and E5M2 (5-bit exponent and 2-bit mantissa). E5M3 follows IEEE 754 conventions. FP4~\cite{rouhani2023microscaling} is of E2M1. For quantization to FP4, we utilize QPyTorch~\cite{zhang2019qpytorch} for simulation.

\section{Additional Experiments}~\label{app:exp}

\subsection{C4 and WikiText}\label{app:c4-wiki-exp}

The C4 and WikiText perplexity of FP16-weight LLaMA models is given in Table~\ref{tab:c4-wiki-fp16} while the INT8-weight version is shown in Table~\ref{tab:c4-wiki-int8}. Both communication volumes are quantized with Flash communication.

\begin{table}[ht]
\caption{LLaMA models' perplexity of C4 (upper rows) and WikiText2 (lower rows) with fine-grained communication quantization with a group size of 128.}
\label{tab:c4-wiki-fp16}
\begin{center}
\begin{small}
\begin{sc}
\begin{tabular}{lHcccHc}
\toprule
Model &	INT8	&	INT8$_{Asym}$	&	FP8	&	INT6$_{Asym}$ &	INT4	&	INT4$_{Asym}$ \\
\midrule
3-8B	&	8.90	&	8.89	&	8.90	&	9.20	&	10.51	&	9.68\\
3-70B	&	6.74	&	6.74	&	6.75	&	6.85	&	7.30	&	7.04\\
2-7B	&	6.98	&	6.98	&	6.98	&	7.07	&	7.50	&	7.21\\
2-13B	&	6.47	&	6.47	&	6.47	&	6.50	&	6.71	&	6.58\\
2-70B	&	5.52	&	5.52	&	5.52	&	5.55	&	5.69	&	5.60\\
\midrule
3-8B	&	6.15	&	6.14	&	6.15	&	6.37	&	7.22	&	6.70\\
3-70B	&	2.86	&	2.86	&	2.87	&	3.00	&	3.44	&	3.21\\
2-7B	&	5.47	&	5.47	&	5.48	&	5.55	&	5.91	&	5.66\\
2-13B	&	4.88	&	4.88	&	4.89	&	4.93	&	5.08	&	4.99\\
2-70B	&	3.32	&	3.32	&	3.32	&	3.35	&	3.47	&	3.40\\
\bottomrule
\end{tabular}
\end{sc}
\end{small}
\end{center}
\end{table}

\begin{table}[ht]
\caption{The 8-bit LLaMA models' perplexity of C4 (upper rows) and WikiText2 (lower rows) with fine-grained communication quantization with a group size of 128.}
\label{tab:c4-wiki-int8}
\begin{center}
\begin{small}
\begin{sc}
\begin{tabular}{lHcccc}
\toprule
Model	&	FP16	&	INT8$_{Asym}$	&	FP8	&	INT6$_{Asym}$	&	INT4$_{Asym}$\\
\midrule
Llama-2-7B	&	7.00	&	7.00	&	7.01	&	7.10	&	7.24\\
Llama-2-13B	&	6.51	&	6.51	&	6.51	&	6.55	&	6.63\\
Llama-2-70B	&	5.54	&	5.54	&	5.54	&	5.57	&	5.62\\
Llama-3-8B	&	9.00	&	9.01	&	9.02	&	9.33	&	9.85\\
Llama-3-70B	&	6.82	&	6.82	&	6.83	&	6.95	&	7.13\\
\midrule
Llama-2-7B	&	5.50	&	5.50	&	5.51	&	5.59	&	5.69\\
Llama-2-13B	&	4.92	&	4.92	&	4.92	&	4.97	&	5.03\\
Llama-2-70B	&	3.35	&	3.35	&	3.35	&	3.39	&	3.43\\
Llama-3-8B	&	6.24	&	6.25	&	6.26	&	6.49	&	6.84\\
Llama-3-70B	&	2.96	&	2.96	&	2.97	&	3.13	&	3.32\\
\bottomrule
\end{tabular}
\end{sc}
\end{small}
\end{center}
\end{table}


\section{Additional Latency Measurements}~\label{app:latency}

\begin{figure*}[ht]
  \centering
  \subfigure{
  \includegraphics[width=0.45\linewidth]{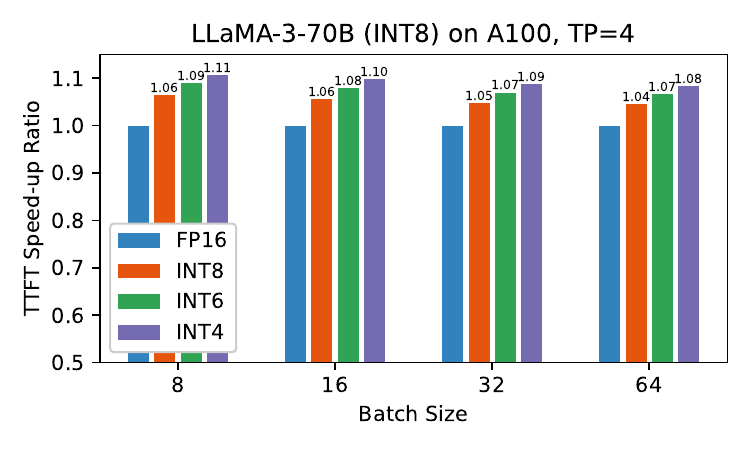}
  }
  \subfigure{
  \includegraphics[width=0.45\linewidth]{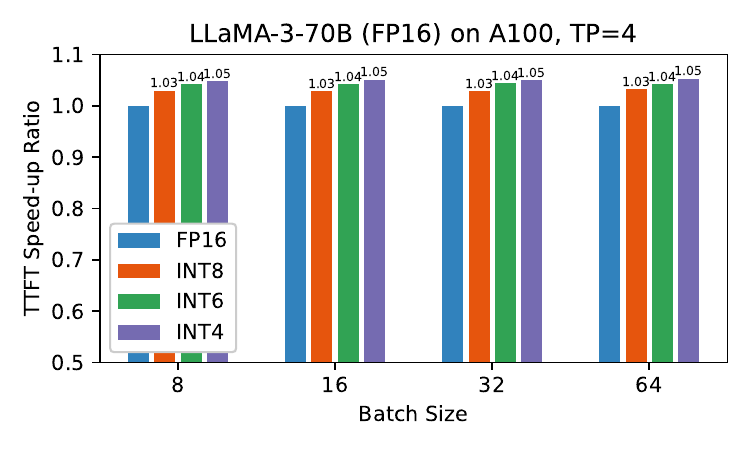}
  }
   \caption{TTFT speed-up ratio of LLaMA-3-70B (INT8 vs. FP16) under various communication quantization bit widths on A100 with TP=4.}
  \label{fig:ttft-comparison-a100-tp4-llama3-72b}
\end{figure*}

We list the latency measurements of LLaMA models under various configurations (weight precision, tensor parallelism, GPU cards) in the following Fig.~\ref{fig:ttft-comparison-l40-tp2-llama3-8b} and Fig.~\ref{fig:ttft-comparison-a100-tp4-llama3-72b}.

\begin{figure*}[ht]
  \centering
  \subfigure{
  \includegraphics[width=0.45\linewidth]{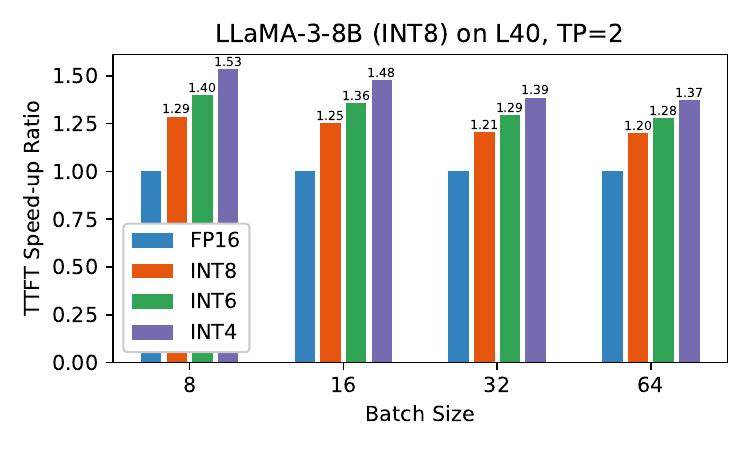}
  }
  \subfigure{
  \includegraphics[width=0.45\linewidth]{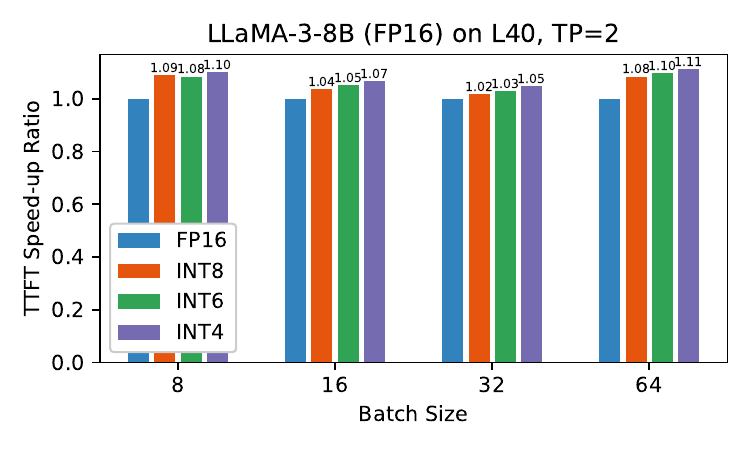}
  }
   \caption{TTFT speed-up ratio of LLaMA-3-8B (INT8 vs. FP16) under various communication quantization bit widths on L40 with TP=2.}
  \label{fig:ttft-comparison-l40-tp2-llama3-8b}
\end{figure*}

\end{document}